\documentclass[11pt,a4paper]{article}
\usepackage[hyperref]{acl2021}
\usepackage{times}
\usepackage{latexsym}
\usepackage{graphicx}
\usepackage{CJKutf8}
\usepackage{amsmath}
\usepackage{bm}
\usepackage{booktabs}
\usepackage{multirow}
\usepackage{url}
\usepackage{breakurl}

\usepackage{tikz-dependency}
\usepackage{amsmath}
\usepackage{graphicx}
\usepackage{pgfplots}
\usepackage{tikz}
\usepackage[center]{subfigure}
\usepackage{tikz-qtree}
\usepgflibrary{arrows.meta}
\usepgfplotslibrary{fillbetween}
\usetikzlibrary{fit, calc, decorations.pathreplacing, positioning, shapes.geometric, backgrounds}
\usepackage{float}

\usepackage{microtype}

\aclfinalcopy 
\def\aclpaperid{2185} 

\title{
An In-depth Study on Internal Structure of Chinese Words
}

\author{
Chen Gong$^{1}$, Saihao Huang$^1$\thanks{$~$ Chen Gong and Saihao Huang make equal contributions to this
work. Zhenghua is the corresponding author.}, Houquan Zhou$^1$, Zhenghua Li$^1$, Min Zhang$^1$,\\ {\bf Zhefeng Wang}$^2$, {\bf Baoxing Huai}$^2$, {\bf Nicholas Jing Yuan}$^2$ \\
$^1$Institute of Artificial Intelligence, School of Computer Science and Technology, \\Soochow University, China; $~~~~~$
$^2$ Huawei Cloud, China  \\
{\tt $^1$\{cgong,shhuang1999,hqzhou\}@stu.suda.edu.cn} \\
{\tt $^1$\{zhli13,minzhang\}@suda.edu.cn} \\
{\tt $^2$\{wangzhefeng, huaibaoxing, nicholas.yuan\}@huawei.com} \\
}

\date{}

\begin{document}
\maketitle
\begin{CJK}{UTF8}{gkai}
\begin{abstract}
Unlike English letters, Chinese characters have rich and specific meanings.
Usually, the meaning of a word can be derived from its constituent characters in some way. 
Several previous works on syntactic parsing propose to annotate shallow word-internal  structures for better utilizing character-level information. 
This work proposes to model the deep internal structures of Chinese words as dependency trees with 11 labels for distinguishing syntactic relationships. 
First, based on newly compiled annotation guidelines, we manually annotate a word-internal structure treebank (WIST) consisting of over 30K multi-char words from Chinese Penn Treebank. 
To guarantee quality, each word is independently annotated by two annotators and inconsistencies are handled by a third senior annotator.  
Second, we present detailed and interesting analysis on WIST to reveal insights on Chinese word formation. 
Third, we propose word-internal structure parsing as a new task, and conduct benchmark experiments using a competitive dependency parser. 
Finally, we present two simple ways to encode word-internal structures, leading to promising gains on  the sentence-level syntactic parsing task.
\end{abstract}
\section{Introduction}\label{sec:intro}







Unlike English, 
Chinese adopts a logographic writing system and contains tens of thousands of distinct characters. 
Many characters, especially frequently used ones, have rich and specific meanings. 

However, words, instead of characters, are often considered as the basic unit in processing Chinese texts. We believe the reason may be two-fold. First, usually a character may have many meanings and usages. Word formation process greatly reduces such char-level ambiguity. Second, by definition, words are the minimal units that express a complete semantic concept or play a grammatical role independently \cite{ctb-xiafei,yu2003ppd}.\footnote{There 
is still a dispute on the word granularity issue \cite{gong-2017-multi,naacl21-latticebert}. Words are defined as a character sequence that is in tight and steady combination. However, the combination intensity is usually yet vaguely qualified according to co-occurrence frequency. We believe this work may also be potentially useful to 
this direction.
}
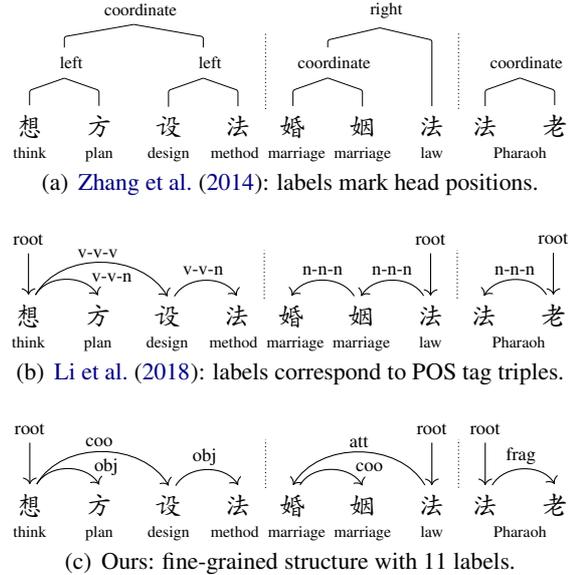
\begin{figure}[tb]
\centering
\subfigure[\citet{zhang-etal-2014-char}: labels mark head positions.]
{
\begin{minipage}[c]{1\linewidth}
    \label{fig:zhangs-tree}
    \centering
    \begin{tikzpicture}[
        scale=0.92,
        level distance=22pt,
        sibling distance=10pt,
        every tree node/.style={align=center, anchor=base},
        frontier/.style={distance from root=50pt},
        postag/.style={text opacity=1, rounded corners=1mm, align=center, font=\scriptsize},
        label/.style={text=white, font=\scriptsize},
        gloss/.style={font=\tiny, anchor=base},
        edge from parent/.style={draw,edge from parent path={(\tikzparentnode.south) {[rounded corners=0.8pt]-- ($(\tikzchildnode |- \tikzparentnode.south) + (0, -5pt)$) -- (\tikzchildnode)}}}
        ]
        \begin{scope}[xshift=-100pt]
            \Tree
            [.\node[postag]{coordinate};
                [. \node[postag]{left}; \node(0_0){想}; \node(0_1){方}; ]
                [. \node[postag]{left}; \node(0_2){设}; \node(0_3){法}; ] ];
            \node[label] at ($(0_0) + (+0.0, +1.1)$) {root};
            \node[gloss] at ($(0_0.base) + (0.0, -0.3)$) {think};
            \node[gloss] at ($(0_1.base) + (0.0, -0.3)$) {plan};
            \node[gloss] at ($(0_2.base) + (0.0, -0.3)$) {design};
            \node[gloss] at ($(0_3.base) + (-0.05, -0.3)$) {method};
        \end{scope}

        \Tree
        [.\node[postag]{right};
            [. \node[postag]{coordinate}; \node(1_0){婚}; \node(1_1){姻}; ]
            \node(1_2){法}; ];
        \node[gloss] at ($(1_0.base) + (0.05, -0.3)$) {marriage};
        \node[gloss] at ($(1_1.base) + (0.0, -0.3)$) {marriage};
        \node[gloss] at ($(1_2.base) + (0.0, -0.3)$) {law};
        
        \begin{scope}[xshift=+57pt]
            \Tree [.\node[postag, text=white]{coordinate}; \edge[draw=none]; [.\node[postag]{coordinate}; \node(2_0){法}; \node(2_1){老}; ] ];
            \node[gloss] at ($(2_0.base)!0.5!(2_1.base) + (0.0, -0.3)$) {Pharaoh};
        \end{scope}
        
        \draw[densely dotted] ($(0_3)!0.5!(1_0) + (0, 0.25)$) to ($(0_3)!0.5!(1_0) + (0, 1.35)$);
        \draw[densely dotted] ($(1_2)!0.5!(2_0) + (0, 0.25)$) to ($(1_2)!0.5!(2_0) + (0, 1.35)$);

    \end{tikzpicture}
    \end{minipage}
}\\
\subfigure[\citet{li-etal-aaai-2018-zhaohai}: labels correspond to POS tag triples.]
{
\begin{minipage}[c]{1\linewidth}
\label{fig:zhaos-tree}
    \centering
    \begin{tikzpicture}[
        scale=0.92,
        level distance=10pt,
        sibling distance=10pt,
        every tree node/.style={align=center,anchor=base},
        frontier/.style={distance from root=20pt},
        postag/.style={fill=none, text=white, rounded corners=1mm, align=center},
        label/.style={font=\scriptsize},
        gloss/.style={font=\tiny, anchor=base},
        edge from parent/.style={draw=none,edge from parent path={(\tikzparentnode.south) {[rounded corners=0.8pt]-- ($(\tikzchildnode |- \tikzparentnode.south) + (0, -5pt)$) -- (\tikzchildnode)}}}
        ]
        \begin{scope}[xshift=-100pt]
            \Tree
            [ \node(0_0){想}; \node(0_1){方}; \node(0_2){设}; \node(0_3){法}; ];
            \draw[<-] (0_1.north) to[out=120, in=60] ($(0_0.north) + (+0.1, 0) $);
            \draw[<-] (0_2.north) to[out=120, in=60] ($(0_0.north) + (+0.1, 0) $);
            \draw[<-] (0_0.north) to ($(0_0) + (0, +0.9)$);
            \draw[<-] (0_3.north) to[out=120, in=60] ($(0_2.north) + (+0.1, 0) $);
            \node[label] at ($(0_0) + (+0.0, +1.1)$) {root};
            \node[label] at ($(0_1) + (+0.2, +0.55)$) {v-v-n};
            \node[label] at ($(0_1) + (+0.0, +0.9)$) {v-v-v};
            \node[label] at ($(0_3) + (-0.48, +0.65)$) {v-v-n};
            \node[gloss] at ($(0_0.base) + (0.0, -0.3)$) {think};
            \node[gloss] at ($(0_1.base) + (0.0, -0.3)$) {plan};
            \node[gloss] at ($(0_2.base) + (0.0, -0.3)$) {design};
            \node[gloss] at ($(0_3.base) + (-0.05, -0.3)$) {method};
        \end{scope}

        \Tree
        [ [\node(1_0){婚}; \node(1_1){姻}; ] \node(1_2){法}; ];
        \draw[<-] (1_0.north) to[out=60, in=120] ($(1_1.north) + (-0.1, 0) $);
        \draw[<-] (1_1.north) to[out=60, in=120] ($(1_2.north) + (-0.1, 0) $);
        \draw[<-] (1_2.north) to ($(1_2) + (0, +0.9)$);
        \node[label] at ($(1_1) + (-0.55, +0.65)$) {n-n-n};
        \node[label] at ($(1_2) + (-0.55, +0.65)$) {n-n-n};
        \node[label] at ($(1_2) + (+0.0, +1.1)$) {root};
        \node[gloss] at ($(1_0.base) + (0.05, -0.3)$) {marriage};
        \node[gloss] at ($(1_1.base) + (0.0, -0.3)$) {marriage};
        \node[gloss] at ($(1_2.base) + (0.0, -0.3)$) {law};
        
        \begin{scope}[xshift=+57pt]
            \Tree [ \node(2_0){法}; \node(2_1){老}; ];
            \draw[<-] (2_0.north) to[out=60, in=120] ($(2_1.north) + (-0.1, 0) $);
            \draw[<-] (2_1.north) to ($(2_1) + (0, +0.9)$);
            \node[label] at ($(2_1) + (-0.55, +0.65)$) {n-n-n};
            \node[label] at ($(2_1) + (+0.0, +1.1)$) {root};
            \node[gloss] at ($(2_0.base)!0.5!(2_1.base) + (0.0, -0.3)$) {Pharaoh}; 
        \end{scope}

        \draw[densely dotted] ($(0_3)!0.5!(1_0) + (0, 0.25)$) to ($(0_3)!0.5!(1_0) + (0, 0.95)$);
        \draw[densely dotted] ($(1_2)!0.5!(2_0) + (0, 0.25)$) to ($(1_2)!0.5!(2_0) + (0, 0.95)$);
        
    \end{tikzpicture}
    \end{minipage}
}\\
\subfigure[Ours: fine-grained structure with 11 labels.]
{
\begin{minipage}[c]{1\linewidth}

\label{fig:ours-tree}
    \centering
    \begin{tikzpicture}[
        scale=0.92,
        level distance=10pt,
        sibling distance=10pt,
        every tree node/.style={align=center,anchor=base},
        frontier/.style={distance from root=20pt},
        postag/.style={fill=none, text=white, rounded corners=1mm, align=center},
        label/.style={font=\scriptsize},
        gloss/.style={font=\tiny, anchor=base},
        edge from parent/.style={draw=none,edge from parent path={(\tikzparentnode.south) {[rounded corners=0.8pt]-- ($(\tikzchildnode |- \tikzparentnode.south) + (0, -5pt)$) -- (\tikzchildnode)}}}
        ]
        \begin{scope}[xshift=-100pt]
            \Tree
            [ \node(0_0){想}; \node(0_1){方}; \node(0_2){设}; \node(0_3){法}; ];
            \draw[<-] (0_1.north) to[out=120, in=60] ($(0_0.north) + (+0.1, 0) $);
            \draw[<-] (0_2.north) to[out=120, in=60] ($(0_0.north) + (+0.1, 0) $);
            \draw[<-] (0_0.north) to ($(0_0) + (0, +0.9)$);
            \draw[<-] (0_3.north) to[out=120, in=60] ($(0_2.north) + (+0.1, 0) $);
            \node[label] at ($(0_0) + (+0.0, +1.1)$) {root};
            \node[label] at ($(0_1) + (+0.1, +0.55)$) {obj};
            \node[label] at ($(0_1) + (+0.0, +0.90)$) {coo};
            \node[label] at ($(0_3) + (-0.48, +0.7)$) {obj};
            \node[gloss] at ($(0_0.base) + (0.0, -0.3)$) {think};
            \node[gloss] at ($(0_1.base) + (0.0, -0.3)$) {plan};
            \node[gloss] at ($(0_2.base) + (0.0, -0.3)$) {design};
            \node[gloss] at ($(0_3.base) + (-0.05, -0.3)$) {method};
        \end{scope}

        \Tree
        [ [ \node(1_0){婚}; \node(1_1){姻}; ] \node(1_2){法}; ];
        \draw[<-] (1_0.north) to[out=60, in=120] ($(1_2.north) + (-0.1, 0) $);
        \draw[<-] (1_1.north) to[out=120, in=60] ($(1_0.north) + (+0.1, 0) $);
        \draw[<-] (1_2.north) to ($(1_2) + (0, +0.9)$);
        \node[label] at ($(1_1) + (+0.1, +0.55)$) {coo};
        \node[label] at ($(1_1) + (-0.05, +0.91)$) {att};
        \node[label] at ($(1_2) + (+0.0, +1.1)$) {root};
        \node[gloss] at ($(1_0.base) + (0.05, -0.3)$) {marriage};
        \node[gloss] at ($(1_1.base) + (0.0, -0.3)$) {marriage};
        \node[gloss] at ($(1_2.base) + (0.0, -0.3)$) {law};

        \begin{scope}[xshift=+57pt]
            \Tree [ \node(2_0){法}; \node(2_1){老}; ];
            \draw[<-] (2_1.north) to[out=120, in=60] ($(2_0.north) + (+0.1, 0) $);
            \draw[<-] (2_0.north) to ($(2_0) + (0, +0.9)$);
            \node[label] at ($(2_1) + (-0.48, +0.70)$) {frag};
            \node[label] at ($(2_0) + (+0.0, +1.1)$) {root};
            \node[gloss] at ($(2_0.base)!0.5!(2_1.base) + (0.0, -0.3)$) {Pharaoh}; 
        \end{scope}
        
        \draw[densely dotted] ($(0_3)!0.5!(1_0) + (0, 0.25)$) to ($(0_3)!0.5!(1_0) + (0, 0.95)$);
        \draw[densely dotted] ($(1_2)!0.5!(2_0) + (0, 0.25)$) to ($(1_2)!0.5!(2_0) + (0, 0.95)$);
        
    \end{tikzpicture}
    \end{minipage}
}
\caption{
Three example words with internal structure under different annotation paradigms. ``想(think of) 方(plan) 设(design) 法(method)'' is a verb and means ``find ways or means to do''. ``婚(marriage) 姻(marriage) 法(law)'' is a noun. ``法老'' is phonetic transliteration of ``Pharaoh''. The three words all contain the character ``法'' under different meanings. 
}
\label{fig:example}

\end{figure}

Roles played by characters in word formation can be divided into three types. 
\textbf{\footnotesize{(1)}} There is a stable and important set of 
\emph{single-char words}, such as ``你'' (you)'', ``的'' (of), and most punctuation marks. 
\textbf{\footnotesize{(2)}} A character having no specific meaning acts as a \emph{part of a single-morpheme word}, such as ``仿佛'' (like) and 
``法(f\v{a})老(l\v{a}o)'' (Pharaoh, 
transliteration of foreign words). 
\textbf{\footnotesize{(3)}} A character corresponds to a \emph{morpheme}, the smallest meaningful unit in a language, and composes a polysyllabic word with other characters. This work  targets multi-char words, and  is particularly interested in the third type which most characters belong to. 

Intuitively, modeling how multiple characters form a word, i.e., the word-formation process, allows us to  more effectively represent the meaning of a word via composing the meanings of characters. 
This is especially helpful for handling rare words,  considering that the vocabulary size of characters is much  smaller than that of words.
In fact, many NLP researchers have tried to utilize char-level word-internal structures for better Chinese understanding. 
Most related to ours, previous studies on syntactic parsing have proposed to annotate word-internal structures to alleviate the data sparseness problem \cite{zhang-etal-2014-char,li-etal-aaai-2018-zhaohai}. However, their annotations mainly consider flat and shallow word-internal structure, as shown in Figure \ref{fig:example}-(a) and (b). 
Meanwhile, researchers try to make use of character information to learn better word embeddings \cite{chen-ijcai2015-joint-char-word-emb,xu-naacl-2016-internal-structure}. 
Without explicitly capturing word-internal structures, these studies have to treat a word as a bag of characters.
See Section \ref{sec:related-work} for more discussion. 

This paper presents an in-depth study on char-level internal structure of Chinese  words. We endeavour to address three questions. \textbf{\footnotesize{(1)}} What are 
the word-formation patterns for Chinese words? \textbf{\footnotesize{(2)}} Can we train a model to predict deep word-internal structures? 
\textbf{\footnotesize{(3)}} Is modeling word-internal structures beneficial for word representation learning? 

For the first question, we propose to use labeled dependency trees to represent word-internal structures, and employ 11 labels to distinguish syntactic roles in word formation. We compile annotation guidelines following the famous textbook of \citet{grammar-notes-zhu-1982} on Chinese syntax, and annotate a high-quality word-internal structure treebank (WIST), consisting of 30K words from Penn Chinese Treebank (CTB) \cite{ctb-xiafei}. We conduct detailed analysis on WIST to gain insights on Chinese word-formation patterns.

For the second question, we propose word-internal structure parsing as a new task, and present benchmark experimental results using a competitive open-source dependency parser. 

For the third question, we investigate two simple ways to encode word-internal structure, i.e., LabelCharLSTM and LabelGCN, and show that using the resulting word representation leads to promising gains on the dependency parsing task.

We release WIST at \url{https://github.com/SUDA-LA/ACL2021-wist}, and also provide a demo to parse the internal structure of any input word.

\section{Related Work}\label{sec:related-work}

\renewcommand\arraystretch{0.8}
\setlength{\tabcolsep}{3pt}
\begin{table*}[!htbp]
\begin{center}
\newcommand{\tabincell}[2]{\begin{tabular}{@{}#1@{}}#2\end{tabular}}
\begin{tabular}{l  l  l  l  }
    \toprule
    Label $~~~~~~~$  & Meaning & Example &  Annotation \\
    \hline
        root & word root & 登场 (come on stage) & \$ $\xrightarrow{\textsf{root}}$ 登 (come) $\xrightarrow{\textsf{obj}}$ 场 (stage)\\
    subj & subject & 年轻 (young) & 年 (age) $\xleftarrow{\textsf{subj}}$ 轻 (small)\\ 
    obj & object & 下雨 (rain) & 下 (drop) $\xrightarrow{\textsf{obj}}$ 雨 (rain)\\ 
    att & attribute modifier & 大衣 (overcoat) & 大 (large) $\xleftarrow{\textsf{att}}$ 衣 (coat) \\
    adv & adverbial modifier & 不同 (different) & 不 (not) $\xleftarrow{\textsf{adv}}$ 同 (same) \\
    cmp & complement modifier & 放下 (put down) & 放 (put) $\xrightarrow{\textsf{cmp}}$下 (down) \\
    coo & coordination & 上下文 (context) & 上 (above) $\xrightarrow{\textsf{coo}}$ 下(below) \\
    pobj & preposition object & 到期 (expire) & 到 (reach) $\xrightarrow{\textsf{pobj}}$期 (deadline) \\
    adjct & adjunct & 走过 (pass by)& 走 (walk) $\xrightarrow{\textsf{adjct}}$过 (by) \\
    frag & fragment & 沙发 (sofa) & 沙 (sand) $\xrightarrow{\textsf{frag}}$发 (send) \\
    repet & repetition & 常常 (often) & 常 (often) $\xrightarrow{\textsf{repet}}$常 (often) \\
\bottomrule   
\end{tabular}
\end{center}
\caption{{The 11 labels adopted in our guidelines for distinguishing syntactic roles in word formation.}}\label{tbl:summary-relation}
\end{table*}
\renewcommand\arraystretch{1.}




\paragraph{Annotating word-internal structure.}

In the deep learning (DL) era, pretraining techniques are extremely powerful in handling large-scale unlabeled data, including Skip-Gram or CBOW models \cite{mikolov2013efficient} for learning context-independent word embedding in the beginning, and the recent ELMo \cite{peters2018deep} or BERT \cite{devlin-2019-bert} for learning context-aware word representations. 
Conversely, in the pre-DL era, there exist few (if any) effective methods for utilizing unlabeled data, and statistical models rely on discrete one-hot features, leading to severe data sparseness for many NLP tasks. This directly motivates annotation of word-internal structure, especially for dealing with rare words. 

Annotation of shallow internal structure of Chinese words was first mentioned in \citet{zhao-2009}, largely based on heuristic rules. 
\citet{li-2011,li-zhou-2012} found that many multi-char words could be divided into two subwords, i.e., root and affix. They annotated structures of about 19K words (35\% of 54,214) in CTB6. Their experiments showed that subword-level syntactic parsing is superior to word-level parsing. 
For the three words in Figure \ref{fig:example}, their approach is only applicable to the second word, i.e., ``婚姻/法''.
As an extension to \citet{li-zhou-2012}, \citet{zhang-etal-2013-char, zhang-etal-2014-char} proposed char-level syntactic parsing by further dividing  subwords into chars. 
As shown in Figure \ref{fig:example}-(a), for each word, they annotated a binary hierarchical tree, using  constituent labels to mark which child constituent is more syntactically important, i.e., left, right, or coordinate. 
In such way, they could convert a word-level constituent/dependency tree into a char-level one.
Similar to \citet{li-zhou-2012}, \citet{cheng2014parsing} annotated internal structure of synthesis (multi-morpheme) words with four relations, i.e., branching,
coordinate, beginning and other parts of a single-morpheme word.






In the DL era, three works have 
studied 
word-internal structure. 
Similarly to our work, \citet{li-etal-aaai-2018-zhaohai} employed dependency trees to encode word-internal structure. 
As shown in Figure \ref{fig:example}-(b), for each multi-char word, they first annotate the part-of-speech (POS) tag of each character, and then determine an unlabeled dependency tree, and finally use a POS tag triple as arc label, corresponding to the POS tags of the modifier/head characters and the whole word.  
However, we argue POS tag triples 
are only loosely related with word-formation patterns, 
not to mention the severe difficulty of annotating 
char-level POS tags in each word.

Recently, \citet{linqian2020} extended \citet{zhang-etal-2014-char} by using an extra label for marking single-morpheme 
words, and annotated hierarchical internal structure of 53K words from a Chinese-English machine translation (MT) dataset. 
\citet{annotate-syntactic-Depling-2019} annotated the internal structure of words with 4 dependency relations. 


In summary, we can see that most previous studies adopted quite shallow hierarchical structure. In contrast, 
this work 
presents a more in-depth investigation on internal structure of Chinese words and employs 11 labels to distinguish different syntactic roles in word formation, as shown in Figure \ref{fig:example}-(c).



\paragraph{Leveraging character information for better word representation.} 
It has already become a standard way in many NLP tasks to obtain char-aware word representation by applying LSTM or CNN to the character sequence of a word, and concatenate it with word embedding as input, such as named entity recognition \cite{chiu2016named}, dependency parsing \cite{zhang-etal-2020-dep}, and constituent parsing \cite{gaddy2018s}.

Another research direction is to leverage character information to obtain better word embeddings. 
\citet{chen-ijcai2015-joint-char-word-emb} extended the CBOW model and proposed to jointly learn character and word embeddings.
Based on \citet{chen-ijcai2015-joint-char-word-emb}, \citet{xu-emnlp-2017-short} proposed to jointly learn embeddings of words, characters, and sub-characters.\footnote{Following this direction, studies tried to explore more character information for better Chinese word representation, such as strokes \cite{cao-aaai-2018-strokes} and ideographic shape \cite{sun-naacl-2019-visual}.}
However, both studies assume that characters contribute equally to the meaning of a word and directly average embeddings of all characters. 
To address this,  
\citet{xu-naacl-2016-internal-structure} extended \citet{chen-ijcai2015-joint-char-word-emb} and proposed a cross-lingual approach to distinguish contribution of characters for a word.
The idea is to translate Chinese words and characters into English words, and use similarities between corresponding English word embeddings for contribution measurement. 
Instead of treating a word as a bag of characters, we experiment with two simple ways to obtain structure-aware word representations. Meanwhile, enhancing their approach with explicit  word-internal structure could be also very interesting.

\paragraph{Utilizing word-internal structure.} 
Word-internal structure have been explored in various NLP tasks.
Several works propose to learn word-internal structure, word segmentation, POS tagging and parsing jointly \cite{zhang-etal-2013-char,zhang-etal-2014-char,li-etal-aaai-2018-zhaohai}, demonstrating the effectiveness of word-internal structure in helping downstream tasks.
\citet{cheng2015synthetic} attempt to convert words
into fine-grained subwords according to the internal structure of words for better dealing with unknown words during word segmentation. 
\citet{linqian2020} propose to integrate the representation of word-internal structure into the input of neural machine translation model, leading to improved translation performance.

\section{Word-internal Structure Annotation}\label{sec:data-annotation}

In this section, we describe in detail the annotation process of WIST. 
As shown in Figure \ref{fig:example}-(c), we adopt dependency trees for representing word-internal structure. 
The reason is two-fold. First, word-formation process correlates with syntax in different ways depending on language type \cite{Aikhenvald-2007-typological}. Such correlation is especially close for Chinese due to its lack of morphological inflections. In particular, \citet{grammar-notes-zhu-1982} presented thorough  investigation on Chinese word formation mainly from a syntactic view. 
Second, as a grammar formalism, dependency tree structure has been widely adopted for capturing sentence-level syntax due to its simplicity and flexibility in representing  relations. Meanwhile, its computational modeling is also developed quite well. 

\paragraph{Annotation guidelines.} 
After several months' survey,  we have compiled  systematic and detailed
 guidelines for word-internal structure annotation. 
Our guidelines are mainly based on the famous textbook on Chinese grammar of \citet{grammar-notes-zhu-1982}. 
We intensively studied all previous works on word-internal structure annotation, which are discussed in Section \ref{sec:related-work}.
We also find that it is quite beneficial to be familiar with guidelines developed by previous annotation projects for Chinese word segmentation \cite{ctb-xiafei,yu2003ppd}.



Our guidelines contain 11 relations specifically designed to capture the internal dependency syntax for Chinese words, as shown in Table \ref{tbl:summary-relation}. 
We derive most of the dependency relations by referring to guidelines of three popular Chinese dependency treebanks, i.e., UD, Harbin Institute Technology Chinese Dependency Treebank (HIT-CDT) \cite{HIT-CDT}, and Chinese Open Dependency Treebank (CODT) \cite{li-etal-2019-semi-supervised}. 
We give very detailed illustrations with examples in our 30-page guidelines to ensure annotation consistency and quality. 
Our guidelines are also gradually improved according to the feedback from the annotators. 

\paragraph{Quality control.} 
We employ 18 undergraduate students as part-time annotators who are familiar with Chinese syntax, and select 6 capable annotators with a lot of data annotation experience as expert annotators 
to handle inconsistent submissions. 
All the annotators (including expert annotators) were paid for their work . 
The salary is determined by both quantity and quality. Besides, we give extra bonus to the annotators with high accuracy. The average salary of the annotators is 30 RMB per hour.
All annotators are trained 
for several hours to be familiar with our guidelines and the usage of annotation tool. 

We apply strict double annotation in order to guarantee quality. 
Each word 
is randomly assigned to two annotators. Two identical submissions are directly used as the final answer. Otherwise, a third expert annotator is asked to decide the final answer after analyzing the two inconsistent annotations. 

\paragraph{Annotation tool.} 
We build a browser-based annotation tool to support the annotation workflow and facilitate project management.

Given an annotation task, all its POS tags \footnote{In CTB5, a word may be annotated with different POS tags under different contexts. For example, ``发展 (development)'' is annotated as NN (noun) in the context ``促进经济发展 (boost the economic development )'', whereas ``发展 (develop)'' is annotated as (VV) verb in the context ``稳定地发展 (develop steadily)''. Therefore, when annotating the word ``发展 (develop/development )'', we present both ``noun'' and ``verb'' to the annotators for reference.''} of the focused word in CTB5 are presented to the annotator, in order to explore multiple internal structures for one word. 
In that case, the annotator can click a checkbox to inform us for further process. 
Please note that the manually annotated POS tags in CTB5 are converted into Universal Dependencies (UD) \footnote{\scriptsize{\url{universaldependencies.org/u/pos/}}} POS tags based on predefined mapping rules, since the original CTB5 POS tags are too fine-grained (33 tags) and difficult for annotators to understand.
The interface also presents several example sentences to improve annotation efficiency. We strongly encourage annotators to look up difficult words or characters in electronic dictionaries.\footnote{Eg., \url{hanyu.baidu.com}; \url{xh.5156edu.com/}} 

\setlength{\tabcolsep}{3pt}
\begin{table}[tb]
\begin{center}
\newcommand{\tabincell}[2]{\begin{tabular}{@{}#1@{}}#2\end{tabular}}
\begin{tabular}{l |r | r  r  r   r }
    \toprule
     & Total \# & 1 & 2 &  3 & $\ge$4 \\
    \hline
    word type & 37,449 &  5.6 & 58.3 & 22.8 & 13.3 \\
    word token & 508,764 & 48.0  & 44.1 & 6.0 & 1.9 \\
    \bottomrule
\end{tabular}
\end{center}
\caption{{Word distr. regarding char number in CTB5. }}\label{tbl:summary-different-num-word}
\end{table}

\paragraph{Data selection.}

Following previous works, we select multi-char words from CTB5 for annotation. 
Table \ref{tbl:summary-different-num-word} shows word distribution  regarding character numbers. 
We can see that only 5.6\% of words in the vocabulary contain one char, but they account for nearly half (48\%) token occurrences in the text. 
The percent of words with two characters is high in both vocabulary (58.3) and text (44.1). 
We discard words containing special symbols such as English letters. Finally, we have annotated 32,954 multi-char words with their internal structure, containing 83,999 dependencies (2.5 characters per word).







\begin{table*}[!tb]
	\addtolength{\tabcolsep}{2pt}
	\begin{center}
			\begin{tabular}{l r r r r r r r r r r r}
			        \toprule
			        &root &att &coo &frag &obj &adv &cmp &adjct &subj &repet &pobj  \\
					\hline  
					Annotation Accuracy &\emph{93.9} &93.1 &88.6 &89.3 &82.6 &80.6 &85.3 &83.5 &62.0 &\textbf{96.0} &48.2 \\
					$~~~~~$ Unlabeled &93.8 &94.2 &92.3 &93.3 &92.7 &88.1 &\emph{97.9} &92.2 &86.9 &\textbf{99.4} &84.5 \\ 
					\hline 
					Parsing Accuracy &\emph{89.0} &\textbf{89.5} &75.8 &80.6 &77.4 &68.0 &84.0 &76.8 &64.2 &81.1 &58.1\\
					$~~~~~$ Unlabeled &89.0 &90.6 &85.4 &84.1 &88.2 &80.7 &\emph{93.5} &80.5 &80.7 &\textbf{97.3} &83.9 \\ 
					\hline  
					 Overall Distribution  &39.2 &\textbf{29.1} &\emph{10.2} &5.7 &5.4 &4.3 &2.3 &1.5 &1.5 &0.6 &0.2  \\
					$~~~~~$Noun (47.2\%) &42.3 &\textbf{33.8} &\emph{11.5} &2.5 &4.4 &2.6 &0.4 &1.1 &1.1 &0.2 &0.1 \\
					$~~~~~$Verb (24.1\%) &42.2 &3.8 &\textbf{17.9} &0.4 &\emph{12.7} &9.6 &7.9 &1.2 &3.1 &0.9 &0.4 \\
					$~~~~~$Proper Noun (13.1\%) &36.6 &\emph{28.4} &2.3 &\textbf{29.6} &0.8 &0.6 &0.1 &0.9 &0.6 &0.3 &0 \\
					$~~~~~$Adjective (7.1\%) &44.4 &\emph{16.5} &\textbf{17.7} &0.7 &7.5 &8.2 &0.6 &0.7 &1.9 &1.6 &0.2 \\
					$~~~~~$Adverb (3.9\%)&45.5 &\textbf{12.1} &10.3 &0.6 &6.4 &\emph{12.1} &1.8 &5.3 &1.0 &2.8 &2.3 \\
					$~~~~~$Numeral (3.7\%)&20.0 &\textbf{75.7} &0.4 &0.1 &0.1 &0.2 &0 &\emph{3.6} &0 &0.1 &0 \\
					$~~~~~$Others (0.9\%)&47.6 &\textbf{15.2} & \emph{8.7} &2.1 &1.4 &7.7 &4.8 &8.2 &0.3 &3.9 &0.1 \\
					\bottomrule
			\end{tabular}
			\caption{Label-wise accuracy and distribution. The first major row presents  annotation accuracy of WIST and  ``unlabeled'' means not considering labels. The second major row gives parsing accuracy on WIST-test, discussed in Section \ref{sec:WIS-parsing}. The third major row gives distribution of different labels for words of different POS tags. }

			\label{table:label distribution}
	\end{center}
\end{table*}

\section{Analysis on Annotated WIST} 

In this section, we analyze the annotated WIST from different aspects in order to gain more insights on Chinese word-formation patterns. 


\paragraph{Inter-annotator consistency.}  
As discussed earlier, 
each word is labeled by two annotators, and inconsistent submissions are handled by a third senior annotator for obtaining a final answer. 
The averaged inter-annotator consistency ratio is 83.0 dependency-wise, i.e., the percent of characters receiving the same head and label from two annotators, and 75.8 word-wise, i.e., the percent of words receiving the same whole trees. 
If we do not consider labels, the unlabeled consistency ratios increase to 87.5 dependency-wise and 85.1 word-wise. 
Although it may be a factor that most annotators are inexperienced in this new annotation task, 
such low consistency ratios indicate that annotating word-internal structure is quite challenging, especially when it comes to distinguishing syntactic roles. 
Meanwhile, this also demonstrates the importance of strict
double annotation, considering that nearly a quarter of words are inconsistent and require handling by senior annotators.  


\paragraph{Annotation accuracy.}

We calculate annotation accuracy by comparing all submissions (as denominator) from annotators against the final answers in WIST. 
Please note that each word is double annotated. 
The overall dependency-wise accuracy for all annotators is 90.9, and word-wise is 86.9. 
If not considering labels, the overall unlabeled accuracy increases to 93.4 and 92.1, dependency- and word-wise respectively. 

The first major row in Table \ref{table:label distribution} shows the label-wise annotation accuracy. 
We divide characters in WIST into 11 groups according to their final-answer labels, and then  calculate the percent of correct submissions for each group. 
The highest accuracy 
is obtained on ``repet'', since its pattern is quite regular. 
Determining the root character also seems relatively easy. 
The lowest accuracy is 62.0 on ``subj'' and 48.2 on ``pobj''.

Comparing unlabeled versus labeled accuracy, the gap is quite large. The extreme case is ``pobj''. Annotators usually can correctly decide the head (84.5\%), but very unlikely choose its true label ``pobj'' (48.2\%). Similarly, accuracy drops by 24.9 for ``subj''. We give more discussions on  annotation difficulties below.



\paragraph{Label distribution.} 
The third major row in Table \ref{table:label distribution} shows distribution of different labels in WIST. 
From the percentage of ``root'' (39.2\%), we can infer that one word contains 2.5 characters on average. 
The overall percent for ``att'' is 29.1, almost half of the remaining labels, meaning that ``att'' appears once every 1.45  words. This reveals that attribute modification is the most dominated pattern in word formation. 
Coordination structure (``coo'') takes the second place with 10.2\%. 
The third most used pattern is fragment (``frag'') with 5.7\%. 
We give more discussion on ``frag'' below. 

Besides the overall distribution, the third major row in Table \ref{table:label distribution} gives label distribution per POS tag.
For clarity, we give the full name of each POS tag (UD, converted from the fine-grained CTB tags) in Table \ref{table:label distribution}, and it means the POS tag of the focused word.
If a word has multiple POS tags, then the same word-internal structure is used for each tag. For example, if a word ``发 (expand) $\xrightarrow{coo}$ 展 (expand)'' has two tags, i.e., Noun and Verb, then the number of ``coo’’ is added by one for both Noun and Verb.
Moreover, a label is repeatedly counted if it appears several times in the same word. 
Due to space limitation, we only present high-frequency POS tags, with percentage shown in parenthesis. 
Please note that we adopt a coarse-grained POS tag set for clarity.

We can see that nouns are mostly formed with ``att'' (33.8\%) and ``coo'' (11.5\%), whereas verbs are with ``coo/obj/adv/cmp'' in the descending order. Proper nouns are evenly dominated by ``frag'' (29.6\%) and ``att'' (28.4\%). It is also obvious that proper nouns tend to be longer, consisting of 2.7 characters according to its ``root'' percentage. 
Numerals are mainly composed via ``att'' (75.7\%) and consist of 5.0 character on average. 

\paragraph{Multiple structures for one word?} 
Many words have multiple meanings.  
Then the question is: how many words really have multiple internal  structures? 
As illustrated in Section \ref{sec:data-annotation}, we show all POS tags to annotators in order to obtain all internal structures of an ambiguous word.
However, in annotated WIST, we find there are only 103 such words with multiple internal structures, accounting for about 0.3\% of all annotated words, 
and 2.7\% of those having multiple POS tags. 
As a typical example, ``制服'' have two structures. 
As a verb, it means ``subdue'' and has ``制(control) $\xrightarrow{cmp}$ 服(tamely)''. As a noun, it means ``uniform'' and has ``制(regulated) $\xleftarrow{att}$ 服(cloth)''. 
This low percentage reveals that most Chinese words actually have very steady internal structure. They have multiple POS tags, mainly because they are used for different syntactic functions without morphological inflections, such as ``发展'' as verb (``develop'') or noun (``development''). 





\paragraph{More on ``frag''.} 
The ``frag'' label is designed to handle all words that have no internal structure due to the lack of semantic composition. 
From Table \ref{table:label distribution}, we can see that ``frag'' accounts for 5.7\% of all labels. 
In order to gain more insights, we collect all 3,528 words containing ``frag'' in WIST, and randomly sample 100 words for investigation. 
Following the brief discussion in Section \ref{sec:intro}, we divide these words into three types, and find that 81 words are proper nouns (such as person name); 16 correspond to transliteration of foreign words; and 3 are single-morpheme words. 

\paragraph{High-order structure distribution.} 
To gain more insights on complex word-formation structure, we focus on all three-char words. 
We find that the root usually lies in the third character by 74.6\%, and the percentage for the second and first characters is only 15.3 and 10.1 respectively. Looking more closely, we find the following four dominated structures. 

\setlength{\tabcolsep}{6pt}
\begin{center}
\newcommand{\tabincell}[2]{\begin{tabular}{@{}#1@{}}#2\end{tabular}}
\begin{tabular}{c r | c r }
    1 $\leftarrow$ 2 $\leftarrow$ 3 & 34.7\% &  
(1 $\rightarrow$ 2) $\leftarrow$ 3 & 34.2\% \\
\hline
1 $\leftarrow$ 2 $\rightarrow$ 3 & 15.3\% &  
1 $\rightarrow$ 2 $\rightarrow$ 3 & 7.0\% \\
\end{tabular}
\end{center}




\paragraph{Difficulties in annotation.} 
Since it is difficult to capture the patterns on unlabeled-dependency inconsistencies, we focus on confusion patterns in label annotation. Among all characters receiving the same head but different labels from two annotators, 20.1\% correspond to ``\{att, adv\}'' confusion due to the ambiguity of the head character being a verb or a noun. 
The second confusion pattern is ``\{coo,frag\}'', with a proportion of 18.6, which are mainly from proper nouns. According to our guidelines, if the meaning of a proper noun is compounding, annotators have to annotate its real internal structures rather than using ``frag''. 
It is also very difficult to distinguish ``obj'' and ``pobj'', since the boundary between prepositions and verbs is vague in Chinese.











\section{Word-internal Structure Parsing}\label{sec:WIS-parsing}


With annotated WIST, we try to address the second question: can we train a model to predict word-internal structure? We adapt the Biaffine parser proposed by \citet{dozat2016deep}, a widely used sentence-level dependency parser, for this purpose, and present results and analysis.  

\subsection{Biaffine Parser}

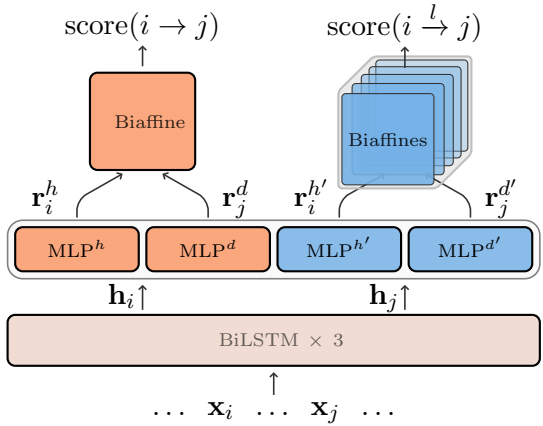
\begin{figure}[tb]
    \centering
    \begin{tikzpicture}[
        connect/.style={
                rounded corners=4pt,
                semithick,
                draw=black!80
            },
        arrow/.style={
                arrows = {-Straight Barb[length=0.5mm]},
                shorten >= 2pt,
                shorten <= 1.5pt,
                thin
            },
        inner arrow/.style={
                arrows = {-Straight Barb[length=0.4mm]},
                shorten >= 2pt,
                shorten <= 2pt,
                thin,
                draw=black!50
            },
        input/.style={
                rectangle,
                rounded corners=1mm,
                thin,
                dashed,
                draw=none,
                minimum width=3.5cm,
                minimum height=0.6cm,
            },
        share/.style={
                minimum height=0.5cm,
                fill={rgb,255:red,230; green,197; blue,180},
                fill opacity=0.6,
                text opacity=1.0,
                draw=black,
                rounded corners=2mm,
            },
        task2/.style={
                minimum height=0.5cm,
                fill={rgb,255:red,251; green,153; blue,104},
                fill opacity=0.6,
                text opacity=1.0,
                draw=black,
                rounded corners=2mm,
            },
        label/.style={
                inner sep=0.5mm,
                fill=white,
                minimum height=0.5cm,
            },
        task1/.style={
                minimum height=0.5cm,
                fill={rgb,255:red,117; green,178; blue,231},
                draw=black,
                rounded corners=2mm
            },
        inner lstm/.style={
                fill=white,
                rectangle,
                rounded corners=1mm,
                semithick,
                draw=black!50,
                fill opacity=0.8
            },
        cell/.style={
                inner sep=2mm,
                rectangle,
                rounded corners=1mm,
                semithick,
                draw=black!50,
            },
        ocell/.style={
                solid,
                minimum height=0.5cm,
                rectangle,
                rounded corners=1mm,
                thick
            },
        dep arrow/.style={
        arrows = {-Latex[round,open,length=8pt,width=6pt]},
        shorten >= 2pt,
        shorten <= 1.5pt,
        thick
        }
        ]
        \centering
        \node [input, inner sep=1pt] [minimum width=4.8cm] (input) at (0, 0) {$\ldots\;\; \mathbf{x}_i \;\;\ldots\;\; \mathbf{x}_j \;\;\ldots$};
        \node [inner sep=0] (EmbedCat) at ($(input.north)$) {};
    
        \node [share, ocell] [minimum width=7cm, minimum height=0.675cm, anchor=south] (lstm) at ($(input.north) + (0, 0.3cm)$) {\scriptsize $\mathrm{BiLSTM}\, \times \, 3$};

        \draw [arrow, connect] ($(EmbedCat.north) + (0cm, -0.15cm)$) -- ($(lstm.south) + (0cm, 0)$);

        \node [share, fill={rgb,255:red,245; green,245; blue,245}, semithick, draw=gray] [minimum width=7cm, minimum height=0.775cm, anchor=south] (share-mlp) at ($(lstm.north) + (0cm, 0.455cm)$) {};

        \node [task2, ocell] [minimum width=1.625cm, minimum height=0.585cm, anchor=south west, fill opacity=0.85] (mlp-1-1-f) at ($(lstm.north west) + (0.1cm, 0.55cm)$) {};
        \node [task2, ocell] [minimum width=1.625cm, minimum height=0.585cm, anchor=south west, fill opacity=0.85] (mlp-1-1-b) at ($(lstm.north west) + (1.825cm, 0.55cm)$) {};
        \node [anchor=base] at  ($(mlp-1-1-f.south) + (0, 0.2cm)$)  {\scriptsize $\mathrm{MLP}^{h}$};
        \node [anchor=base] at  ($(mlp-1-1-b.south) + (0, 0.2cm)$)  {\scriptsize $\mathrm{MLP}^{d}$};
    
        \node [task2, ocell, fill=none, draw=none] [minimum width=1.75cm, minimum height=0.40cm, anchor=south east, fill opacity=0.5, draw opacity=0.6] (mlp-1-2-b) at ($(mlp-1-1-b.north east) + (0, 0.5cm)$) {};
    
        \node [task2, ocell, anchor=south, fill opacity=0.85, minimum height=1.29cm, minimum width=1.29cm, align=center] (arc-biaffine) at ($(mlp-1-1-f.north)!0.5!(mlp-1-1-b.north) + (0, 0.73cm)$) {\scriptsize $\mathrm{Biaffine}$};
    
        \draw [arrow, connect] ($(mlp-1-1-b.north) + (0, 0.06cm)$) -- ($(mlp-1-1-b.north) + (0, 0.35cm)$) -- ($(arc-biaffine.south) + (0.2, 0)$);
        \draw [arrow, connect] ($(mlp-1-1-f.north) + (0, 0.06cm)$) -- ($(mlp-1-1-f.north) + (0, 0.35cm)$) -- ($(arc-biaffine.south) + (-0.2, 0)$);
    
        \node [task1, ocell] [minimum width=1.625cm, minimum height=0.585cm, anchor=south east, fill opacity=0.85] (mlp-2-1-f) at ($(lstm.north east) + (-1.825cm, 0.55cm)$) {};
        \node [task1, ocell] [minimum width=1.625cm, minimum height=0.585cm, anchor=south east, fill opacity=0.85] (mlp-2-1-b) at ($(lstm.north east) + (-0.1cm, 0.55cm)$) {};
        \node [anchor=base] at  ($(mlp-2-1-f.south) + (0, 0.2cm)$)  {\scriptsize $\mathrm{MLP}^{h'}$};
        \node [anchor=base] at  ($(mlp-2-1-b.south) + (0, 0.2cm)$)  {\scriptsize $\mathrm{MLP}^{d'}$};
    
        \node [anchor=south, minimum height=1.29cm, minimum width=1.29cm, draw=none] (label-biaffine-invis) at ($(mlp-2-1-f.north)!0.5!(mlp-2-1-b.north) + (0, 0.73)$) {};
    
        \filldraw[task1, ocell, fill={rgb,255:red,235; green,235; blue,235}, draw=none, rounded corners=0.5mm] ($(mlp-2-1-f.north)!0.5!(mlp-2-1-b.north) + (-0.875, 0.5)$) -- ($(mlp-2-1-f.north)!0.5!(mlp-2-1-b.north) + (0.415, 0.5)$) -- ($(mlp-2-1-f.north)!0.5!(mlp-2-1-b.north) + (0.875, 0.96)$) -- ($(mlp-2-1-f.north)!0.5!(mlp-2-1-b.north) + (0.875, 2.25)$) -- ($(mlp-2-1-f.north)!0.5!(mlp-2-1-b.north) + (-0.415, 2.25)$) -- ($(mlp-2-1-f.north)!0.5!(mlp-2-1-b.north) + (-0.875, 1.79)$) -- cycle;
    
        \draw [arrow, connect] ($(mlp-2-1-b.north) + (0, 0.06cm)$) -- ($(mlp-2-1-b.north) + (0, 0.35cm)$) -- ($(label-biaffine-invis.south) + (0.2, 0)$);
        \draw [arrow, connect] ($(mlp-2-1-f.north) + (0, 0.06cm)$) -- ($(mlp-2-1-f.north) + (0, 0.35cm)$) -- ($(label-biaffine-invis.south) + (-0.2, 0)$);
    
        \filldraw[task1, ocell, fill=none, draw=lightgray, rounded corners=0.5mm] ($(mlp-2-1-f.north)!0.5!(mlp-2-1-b.north) + (-0.875, 0.5)$) -- ($(mlp-2-1-f.north)!0.5!(mlp-2-1-b.north) + (0.415, 0.5)$) -- ($(mlp-2-1-f.north)!0.5!(mlp-2-1-b.north) + (0.875, 0.96)$) -- ($(mlp-2-1-f.north)!0.5!(mlp-2-1-b.north) + (0.875, 2.25)$) -- ($(mlp-2-1-f.north)!0.5!(mlp-2-1-b.north) + (-0.415, 2.25)$) -- ($(mlp-2-1-f.north)!0.5!(mlp-2-1-b.north) + (-0.875, 1.79)$) -- cycle;
    
        \node [anchor=south west, minimum height=1.29cm, minimum width=1.29cm, draw=none] (label-biaffine-f) at ($(mlp-2-1-f.north)!0.5!(mlp-2-1-b.north) + (-0.875, 0.5)$) {};

        \node [anchor=north east, minimum height=1.29cm, minimum width=1.29cm, draw=none] (label-biaffine-b) at ($(mlp-2-1-f.north)!0.5!(mlp-2-1-b.north) + (0.875, 2.25)$) {};  

        \node [fill={rgb,255:red,117; green,178; blue,231}, fill opacity=0.3, minimum height=1.21cm, minimum width=1.21cm, draw=black!80, rounded corners=0.5mm, align=center, inner sep=0.1mm] at ($(label-biaffine-f)!1.0!(label-biaffine-b)$) {};
    
        \node [fill={rgb,255:red,117; green,178; blue,231}, fill opacity=0.4, minimum height=1.21cm, minimum width=1.21cm, draw=black!80, rounded corners=0.5mm, align=center, inner sep=0.1mm] at ($(label-biaffine-f)!0.8!(label-biaffine-b)$) {};
    
        \node [fill={rgb,255:red,117; green,178; blue,231}, fill opacity=0.5, minimum height=1.21cm, minimum width=1.21cm, draw=black!80, rounded corners=0.5mm, align=center, inner sep=0.1mm] at ($(label-biaffine-f)!0.57!(label-biaffine-b)$) {};

        \node [fill={rgb,255:red,117; green,178; blue,231}, fill opacity=0.7, minimum height=1.21cm, minimum width=1.21cm, draw=black!80, rounded corners=0.5mm, align=center, inner sep=0.1mm] at ($(label-biaffine-f)!0.3!(label-biaffine-b)$) {};
        
        \node [fill={rgb,255:red,117; green,178; blue,231}, fill opacity=0.9, minimum height=1.21cm, minimum width=1.21cm, draw=black!80, rounded corners=0.5mm, align=center, inner sep=0.1mm] at ($(label-biaffine-f)!0.0!(label-biaffine-b)$) {\scriptsize $\mathrm{Biaffines}$};
    
        \draw [arrow, connect, rounded corners=1.2pt, shorten >= 2pt] ($(lstm.north) + (-1.725cm, 0)$) -- ($(share-mlp.south) + (-1.725cm, 0)$);
        \draw [arrow, connect, rounded corners=1.2pt, shorten >= 2pt] ($(lstm.north) + (1.725cm, 0)$) -- ($(share-mlp.south) + (1.725cm, 0)$);
        \node[anchor=base] at ($(share-mlp.south) + (-2cm,-0.32cm) $) {$\mathbf{h}_{i}$};
        \node[anchor=base] at ($(share-mlp.south) + (1.45cm,-0.32cm) $) {$\mathbf{h}_{j}$};

        \node[anchor=base] at ($(share-mlp.north) + (-3.0cm,0.18cm) $) {$\mathbf{r}_{i}^{h}$};
        \node[anchor=base] at ($(share-mlp.north) + (-0.5cm,0.18cm) $) {$\mathbf{r}_{j}^{d}$};
        \node[anchor=base] at ($(share-mlp.north) + (0.5cm,0.18cm) $) {$\mathbf{r}_{i}^{h'}$};
        \node[anchor=base] at ($(share-mlp.north) + (3.0cm,0.18cm) $) {$\mathbf{r}_{j}^{d'}$};

        \node[anchor=base] at ($(arc-biaffine.north) + (0cm,0.5cm) $) (arc-biaffine-label) {$\mathrm{score}(i \rightarrow j)$};
        \node[anchor=base] at ($(label-biaffine-invis.north) + (0,0.5cm) $) (label-biaffine-label) {$\mathrm{score}(i \xrightarrow{l} j)$};

        \draw [arrow, connect, rounded corners=1.2pt, shorten >= 2pt] ($(arc-biaffine-label.south) + (0, -0.25cm) $) -- ($(arc-biaffine-label.south) + (0, 0.15cm) $);
        \draw [arrow, connect, rounded corners=1.2pt, shorten >= 2pt, line cap=round] ($(label-biaffine-label.south) + (0, -0.25cm) $) -- ($(label-biaffine-label.south) + (0, 0.15cm) $);
    
    \end{tikzpicture}
    
    \caption{
      The basic architecture of Biaffine Parser. 
    }
    \label{fig:biaffine-parser}
\end{figure}
We adopt the SuPar implementation released by  \citet{zhang-etal-2020-dep}.\footnote{\url{https://github.com/yzhangcs/parser}}  
As a graph-based parser, Biaffine parser casts a tree parsing task as searching for a maximum-scoring tree from a fully-connected graph, with nodes corresponding to characters in our case. As shown in Figure \ref{fig:biaffine-parser}, it adopts standard encoder-decoder architecture, consisting of the following components.  

\textbf{Input layer.} 
Given an input sequence, 
each item is represented as a dense vector $\mathbf{x}_i$. 
For word-internal structure parsing, an item corresponds to a character, and we use char embedding. 
\begin{equation}
\mathbf{x}_i = \mathbf{emb}(c_i) 
\end{equation}

\textbf{BiLSTM encoder.} 
Then, a three-layer BiLSTM is applied to obtain  context-aware representations. 
We denote the hidden vector of the top-layer BiLSTM for  the i-th position as $\mathbf{h}_i$.

\textbf{Biaffine scorer.} 
Two separate MLPs are applied to each $\mathbf{h}_i$, resulting in 
two lower-dimensional vectors $ \mathbf{r}_{i}^{h}$ (as head) and $\mathbf{r}_{i}^{d}$ (as dependent).
Then the score of a dependency $i \rightarrow j$ is obtained via a biaffine attention over $\mathbf{r}_{i}^{h}$ and $\mathbf{r}_{j}^{d}$. 
Scoring of labeled dependencies such as $i \xrightarrow{l} j$ is analogous. 



\textbf{Decoder. }
With the scores of all dependencies, 
we adopt the first-order algorithm of \citet{Eisner2000} 
to find the optimal unlabeled dependency tree, and then independently decide the highest-scoring label for each arc. 

\textbf{Training loss.} 
During training, the parser computes two independent cross-entropy losses for each position, i.e., maximizing the probability of its correct head and
the correct label between them. 

\subsection{Settings}

\textbf{Data.} We randomly split all words in WIST into three parts, 2,500/5,000 as development/test data and remaining as training data. 

\textbf{Hyperparameters.} 
We set the dimension of char embeddings to 100. 
We obtain pre-trained character embeddings by training  word2vec on Chinese Gigaword Third Edition.
In order to see effect of contextualized character representations, we apply BERT \cite{devlin-2019-bert} \footnote{BERT-base-Chinese：\url{https://github.com/google-research/bert}} to each word as a char sequence. The output vectors of the top four layers are concatenated and reduced into a dimension of 100 via an MLP. 
For other hyper-parameters, we keep the default configuration in SuPar.

\textbf{Evaluation metrics.} 
We adopt the standard unlabeled and labeled attachment score (UAS/LAS), i.e., the percent of characters that receives the correct head (and label). 
The complete match (CM) is the percent of words having correct whole trees.

\setlength{\tabcolsep}{3.6pt}
\begin{table}[tb]
\begin{center}
\begin{tabular}{l  rr rrr }
\toprule
\multirow{2}{*}{} 
& \multicolumn{2}{c}{Dev }
& \multicolumn{3}{c}{Test} \\
\cmidrule(lr){2-3} \cmidrule(lr){4-6}
& UAS & LAS & UAS & LAS & CM\\
\hline
Random & 81.18 & 76.15  & 80.63 & 75.58 & 65.13 \\[2pt]
\multirow{2}*{Pretrained}
 & 82.42 & 77.30 & 81.64 & 76.98  & 67.09 \\
 & +1.24 & +1.15 & +1.01 & +1.40 & +1.96\\[2pt]
\multirow{2}*{BERT} 
 & 88.27 & 85.18 & 88.33 & 84.98 &77.72\\
 & +5.85 & +7.88 & +6.69 & +8.00 & +10.63\\
\bottomrule
\end{tabular}
\caption{Results of word-internal structure parsing using different character representations.} 
\label{iwdp-result}
\end{center}
\end{table}

\subsection{Results}

Table \ref{iwdp-result} shows the main results  
under different char representations.  
It is obvious that using randomly initialized char embeddings, the parser can only reach about 76 in LAS. This shows that parsing word-internal structure is very challenging without using extra resources. 
When we pretrain char embeddings on large-scale labeled data, the performance can be consistently improved by over 1 point in both UAS/LAS, and nearly 2 points in CM. 
Finally, employing the contextualized character representations dramatically improves performance further by about 6/8/10 points in UAS/LAS/CM. 

However, even with BERT, model performance still lags behind averaged human performance (90.9 in LAS) by large margin.
Our experienced annotators can even reach more than 94. 
Our experience in manual annotation points out two possible directions to enhance the model: 1) making use of sentence-level contextual information; 2) leveraging the meanings in dictionaries, usually in the form of explanation or example sentences. 
We leave them for future exploration. 



\textbf{Analysis on label-wise accuracy.}
The second major row in Table \ref{table:label distribution} reports accuracy regarding different labels for the model with BERT. 
The model achieves the highest accuracy on ``att'' and ``root'', possibly because the two labels take very large proportion in the data for sufficient model training. 
By contrast, ``pobj'' and ``subj'' have the lowest accuracy, and are difficult for models as well as discussed in Section \ref{sec:data-annotation}. 
This leads to 
another observation that model accuracy is roughly correlated with annotation accuracy, implying the difficulties for human and model are usually consistent. 










\section{Utilizing Word-internal Structure}

This section presents a preliminary study on utilizing word-internal structure, aiming to address the third question: is modeling word-internal structures beneficial for word representation learning? 

We use sentence-level dependency parsing as the focusing task \cite{KublerDEP09}, mainly considering resemblance in tree structure representation and close relatedness between the two tasks. 
Given an input sentence $w_0w_1...w_m$, 
the goal of dependency parsing is to find an optimal dependency tree for the sentence. 
Again, we adopt SuPar \cite{zhang-etal-2020-dep} for implementation of Biaffine parser \cite{dozat2016deep} 
as our basic parser. 

\subsection{Methods}

The basic parser applys a BiLSTM over character sequence to obtain word representation. 
In this part, we propose two  simple alternative methods to encode internal structure shown in Figure \ref{fig:example}-(c).

\textbf{Basic CharLSTM method.} For each word, the basic Biaffine parser uses the concatenation of word embeddings and CharLSTM outputs to represent each word in the input layer:
\begin{equation}\label{eq:charlstm-eq}
\begin{split}
& \mathbf{x}_i = \mathbf{emb}(w_i) 
 \oplus \textup{CharLSTM}(w_i) \\
& \textup{CharLSTM}(w_i) \leftarrow \textup{BiLSTM}(...,\mathbf{z}_{k},...) \\
& \mathbf{z}_{k} = \mathbf{emb}(c_{i,k}) 
\end{split}
\end{equation}
where $c_{i,k}$ is the k-th character of $w_i$. The final word representation from $\textup{CharLSTM}(w_i)$ 
is obtained by concatenating two last-timestamp hidden output vectors of a one-layer BiLSTM. 


\textbf{LabelCharLSTM Method.} 
Considering that the word is usually very short and a bare label itself provides rich syntax information,  
we propose a straightforward extension to CharLSTM, named as LabelCharLSTM, via minor modification.
\begin{equation}\label{eq:labelcharlstm-eq}
\begin{split}
& \mathbf{z}_{k} = \mathbf{emb}(c_{i,k}) \oplus \mathbf{emb}(l_{i,k})
\end{split}
\end{equation}
where $l_{i,k}$ represents the label between $c_{i,k}$ and its head in the word-internal structure. 
\textbf{LabelGCN method.} 
Previous work show that GCN is very effective in encoding
syntactic trees \cite{marcheggiani-titov-2017,zhang-etal-2018}. 
We follow the implementation of \citet{zhang-etal-2018} and use a two-layer GCN as a more sophisticated way. 
In order to utilize labels, we extend vanilla GCN to have the same input with 
LabelCharLSTM, i.e., $\mathbf{z}_{k}$.
We obtain the final word representation by performing average pooling over the output vectors of the top-layer GCN.  


\subsection{Experiments}

\paragraph{Settings.} Following \citet{chen-manning-2014fast}, we conduct experiments on CTB5 with the same data split (16,091/803/1,910 sentences) and constituent-to-dependency conversion. 
Both char/label embeddings are randomly initialized and have the same dimension of 50. 
For the parsers using gold-standard POS tags, we randomly initialized the POS tagging embeddings and set the dimension to 50.
For other hyperparameters, we adopt the default configuration of SuPar, including the pre-trained word embeddings. 

For multi-char words without annotated internal structure, we use the automatic outputs from the trained parser with BERT in Section \ref{sec:WIS-parsing}, so that every word corresponds to a single structure.  

We use word-wise UAS/LAS/CM for evaluation, and punctuation is excluded in all metrics.

\paragraph{Main results.}

Table \ref{table:results-dependency} shows the parsing performance. We can see that both LabelCharLSTM and LabelGCN substantially outperform the basic CharLSTM method. 
LabelGCN achieves the best performance on UAS and LAS, with a gain of 0.71 and 0.80 respectively. 

The fourth row reports performance of LabelGCN without using label embedding, leading to consistent accuracy drop, demonstrating the usefulness of rich labels, which is a key contribution of this work, despite the extra annotation effort. 




\begin{table}[tb]
\setlength{\tabcolsep}{6.5pt}
\centering
\begin{tabular}{lccccc}
    \toprule
    &  UAS & LAS & CM \\
    \hline
    Basic CharLSTM &88.31 &85.96 &32.04 \\
    LabelCharLSTM    &88.78 &86.51 &\textbf{33.19} \\
    \hline
    LabelGCN       &\textbf{89.02}  &\textbf{86.76} & 32.93 \\
    $~~~~~$ w/o label  &88.66 &86.28 &32.20\\
    \bottomrule
\end{tabular}
    \caption{Parsing performance on CTB5-test.}
    \label{table:results-dependency}
\end{table}




\begin{table}[!tb]
\setlength{\tabcolsep}{4.2pt}
\centering
\begin{tabular}{lrrrr}
    \toprule
   & all & $>2$ &$\le2$ & unk.  \\
    \hline
    Basic CharLSTM &85.96& 86.42 &82.03 &81.73   \\[2pt]
    \multirow{2}*{LabelGCN} 
    &86.76& 87.10 &83.79 &84.30  \\
    & +0.80 &+0.68 & +1.76 &+2.57  \\
    \bottomrule
\end{tabular}
    \caption{Parsing LAS regarding to word frequency.}
    \label{table:rare-word-las}
\end{table}

\paragraph{Analysis on rare words.} 
To gain more insights on how word-internal structure helps word representation learning, we divide the words in CTB5-test into several groups according to their frequency in CTB5-train, and report fine-grained accuracy in Table \ref{table:rare-word-las}. 
We can see that the overall performance gain is mostly contributed by improvement over rare words with low frequency or totally unknown. 
This verifies that word-internal structures can help the model to better represent rare words.

\begin{table}[t]
\setlength{\tabcolsep}{2.5pt}
\centering
\begin{tabular}{lrrrrr}
    \toprule
    &  UAS & LAS \\
    \hline
   \citet{ma-2017-neural} &89.05 &87.74 \\
   \citet{dozat2016deep} &89.30 &88.23 \\
   \citet{acl18-ma-stackpointer} &\textbf{90.59}&\textbf{89.29}\\
    \hline
    Basic CharLSTM &91.11 &89.91  \\
    LabelCharLSTM    &\textbf{91.31} &90.15 \\
    LabelGCN       &\textbf{91.31}  &\textbf{90.16} \\
    \bottomrule
\end{tabular}
    \caption{Parsing performance with gold-standard POS tags on CTB5-test.}
    \label{table:results-dependency-goldpos}
\end{table}

\paragraph{Results with gold-standard POS tags.} As suggested by a reviewer, we train our parser with gold-standard POS tags by concatenating the original input (i.e., $\mathbf{x}_{i}$ in Equation \ref{eq:charlstm-eq}) with gold-standard POS tag embeddings, in order to compare with previous works. 
Table \ref{table:results-dependency-goldpos} shows the results. 
Compared with the Basic CharLSTM results in Table \ref{table:results-dependency}, using gold-standard POS tags as extra features for the Basic CharLSTM leads to substantial improvements by 2.80 and 3.95 in UAS and LAS respectively, and outperforms the previous works as presented in Table \ref{table:results-dependency-goldpos}, showing that the basic CharLSTM can be served as a strong baseline model.



Compared with the Basic CharLSTM, utilizing word-internal structure with LabelCharLSTM or LabelGCN achieves consistently better performance by 0.24 and 0.25 respectively in LAS in the scenario of using gold-standard POS tags.
Besides the strong baseline, another reason that the improvement brings by the internal-word structure is slight when using gold-standard POS tags is that a part of linguistic information in the POS tags and the word-internal structures may be overlapping.




\section{Conclusions}

This paper presents a thorough study on internal structures of Chinese words. 
First, 
we annotate a high-quality word-internal structure treebank covering over 30K words in CTB5, named as WIST. 
Second, we perform analysis on WIST from different perspectives and draw many interesting findings on Chinese word-formation patterns. 
Third, we propose word-internal structure as a new task, and present benchmark results using a popular dependency parser. 
Finally, we 
conduct preliminary experiments with two simple methods, i.e., LabelCharLSTM and LabelGCN, to encode word-internal structure as extra word representation, and find promising performance gains on the sentence-level dependency parsing task. 
Analysis shows that the rich dependency labels adopted in WIST play a key role, and word-internal structure is most beneficial for rare word representation. 









\section*{Acknowledgments}
The authors would like to thank the anonymous reviewers for the helpful comments.
We are very grateful to Guodong Zhou for the inspiring discussions and suggestions on Chinese word-internal structures.
We thank Kaihua Lu for building
the annotation system, and Mingyue Zhou, Haoping Yang, and Yahui Liu for their help in compiling annotation guidelines, and all the annotators for their hard work in data annotation. This work is supported by the National Key Research and Development Program of China under Grant No. 2017YFB1002104.

\bibliographystyle{acl_natbib}
\bibliography{acl2021,anthology}


\end{CJK}
\end{document}


\fontsize{12}{15}
\selectfont
\maketitle

\section{Annotation Interface}

This is a brief introduction to our browser-based annotation platform, 
which we have built to better support our long-term data annotation and guarantee annotation quality. 
The web page is \url{http://url}.

\begin{figure}[hptb]
\centering
\includegraphics[width=1.0\textwidth]{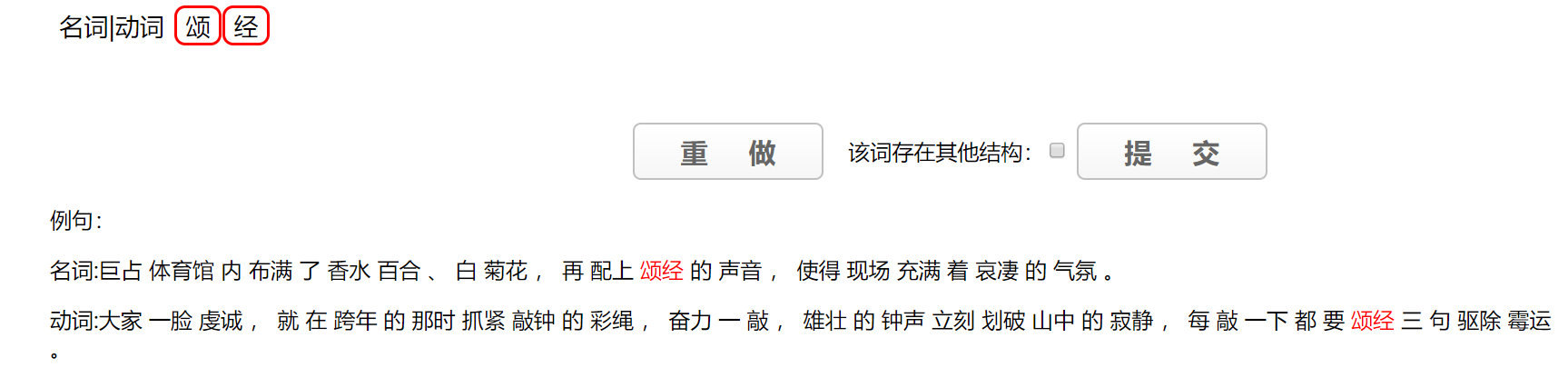}
\caption{
Before annotation: the characters to be annotated are marked with red circles.
}
\label{fig:anno-sys-interface-begin}
\end{figure}

The annotation interface is shown in Figure \ref{fig:anno-sys-interface-begin}. 
At the front of each character, we give all the possible part-of-speech (POS) tags for the annotator to refer to. At the same time, we give the corresponding example sentences for different POS tags in the bottom of the labeling task. Meanwhile, to highlight selected characters to be annotated, the system marks the corresponding characters with red circles.

\begin{figure}[hptb]
\centering
\includegraphics[width=1.0\textwidth]{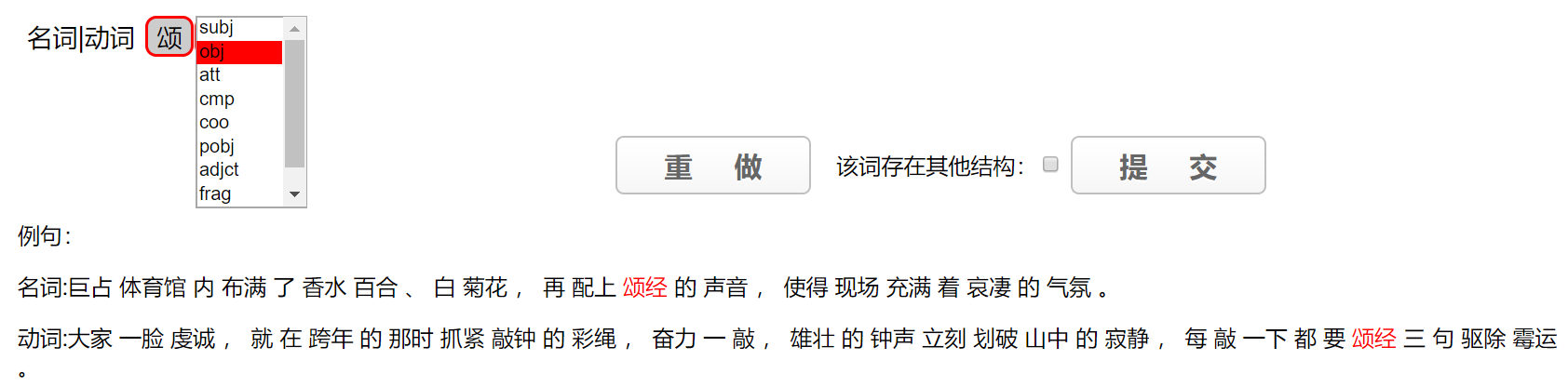}
\caption{
During annotation: the relation label to be selected.
}
\label{fig:anno-sys-interface-select-label}
\end{figure}

\begin{figure}[hptb]
\centering
\includegraphics[width=1.0\textwidth]{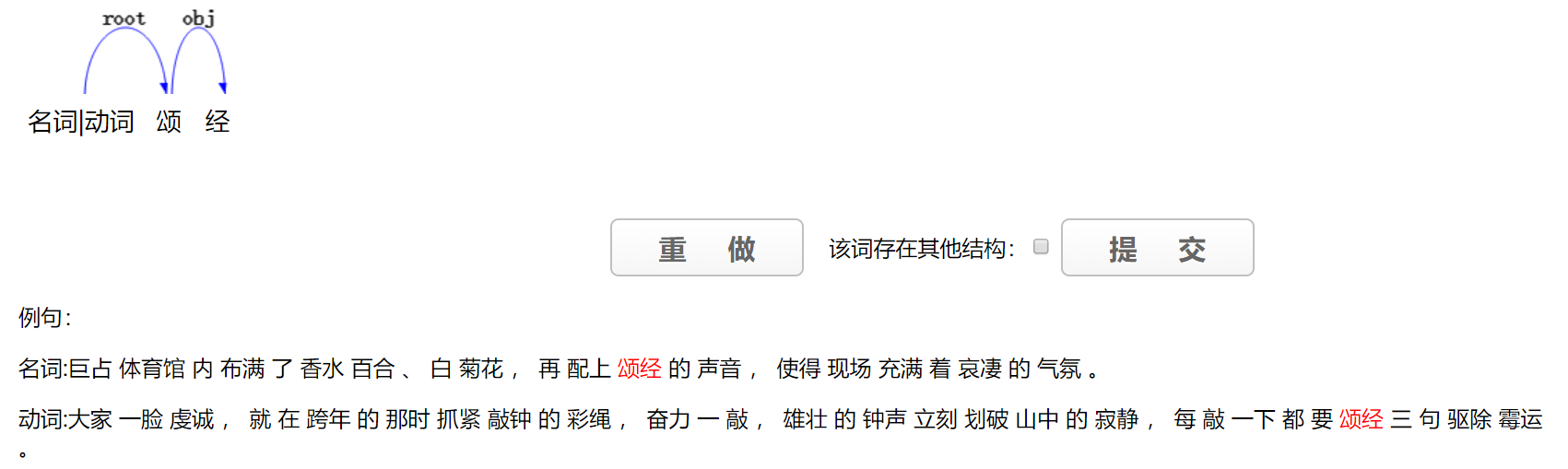}
\caption{
After annotation: the characters are annotated with their heads and red circles disappear.
}
\label{fig:anno-sys-interface-finish}
\end{figure}

To annotate a dependency, annotators just need to first click the head character, and then click the modifier character, and finally select the relation label, as show in Figure \ref{fig:anno-sys-interface-select-label}.

Annotators can only annotate head characters for the selected characters, and must complete all selected tasks before submission. Once the head character is decided, the corresponding red circle disappear, as shown in Figure \ref{fig:anno-sys-interface-finish}.
Annotator may also cancel a dependency by right-click on the modifier character.

We provide a strict annotation guideline, and require annotators to study and follow it. 

The annotated results, before submission, should be a legal tree: 1) single-headed, 2) acyclic, 3) single-root.

\section{Workflow Control}

We design a sophisticated workflow so that the data quality can be guaranteed through the platform and avoid further offline process of the annotated data.

First, the system dynamically assigns each task to two random annotators. 
If the two submissions are not the same (inconsistent dependency or relation label), a third expert annotator will compare them and decide a single answer. 
As a kind of feedback, the platform will return the answer to the annotator who makes mistakes and 
the annotator must correct his/her wrong submission accordingly for the sake of improvement. 
If an annotator insists that the answer is wrong, he/she may click a complaint button 
and the platform then distributes the task to a final senior expert. 
We find this procedure is extremely effective for quality control.